\documentclass[letterpaper,oneside]{article} 
\usepackage{aaai25}  
\usepackage{times}  
\usepackage{helvet}  
\usepackage{courier}  
\usepackage{cuted}
\usepackage[hyphens]{url}  
\usepackage{graphicx} 
\usepackage{natbib}  
\usepackage{caption} 
\usepackage{amsmath}
\usepackage{amssymb}
\usepackage{xcolor}
\usepackage{algorithmic}
\usepackage{multirow}
\usepackage{booktabs}
\usepackage{algorithm}
\usepackage{algorithmic}

\usepackage{newfloat}
\usepackage{listings}
\DeclareCaptionStyle{ruled}{labelfont=normalfont,labelsep=colon,strut=off} 
\floatstyle{ruled}
\newfloat{listing}{tb}{lst}{}
\floatname{listing}{Listing}
\title{SAM-Aware Graph Prompt Reasoning Network for \\Cross-Domain Few-Shot Segmentation}
\author{
    Shi-Feng Peng\textsuperscript{\rm 1}\equalcontrib, 
    Guolei Sun\textsuperscript{\rm 2}\equalcontrib, 
    Yong Li\textsuperscript{\rm 1}, 
    Hongsong Wang\textsuperscript{\rm 3}, 
    Guo-Sen Xie\textsuperscript{\rm 1}\thanks{Corresponding Author.} \\
}
\affiliations{
    \textsuperscript{\rm 1}School of Computer Science and Engineering, Nanjing University of Science and Technology, Nanjing, China\\
    \textsuperscript{\rm 2}Computer Vision Laboratory, ETH Zurich\\
    \textsuperscript{\rm 3}School of Computer Science and Engineering, Southeast University, Nanjing, China\\
    
    psf.nerver@gmail.com; guolei.sun@vision.ee.ethz.ch; 
    yong.li@njust.edu.cn; gsxiehm@gmail.com

}

\begin{document}

\maketitle

\begin{abstract}
The primary challenge of cross-domain few-shot segmentation (CD-FSS) is the domain disparity between the training and inference phases, which can exist in either the input data or the target classes. Previous models struggle to learn feature representations that generalize to various unknown domains from limited training domain samples. In contrast, the large-scale visual model SAM, pre-trained on tens of millions of images from various domains and classes, possesses excellent generalizability. In this work, we propose a SAM-aware graph prompt reasoning network (GPRN) that fully leverages SAM to guide CD-FSS feature representation learning and improve prediction accuracy. Specifically, we propose a SAM-aware prompt initialization module (SPI) to transform the masks generated by SAM into visual prompts enriched with high-level semantic information. Since SAM tends to divide an object into many sub-regions, this may lead to visual prompts representing the same semantic object having inconsistent or fragmented features. We further propose a graph prompt reasoning (GPR) module that constructs a graph among visual prompts to reason about their interrelationships and enable each visual prompt to aggregate information from similar prompts, thus achieving global semantic consistency. Subsequently, each visual prompt embeds its semantic information into the corresponding mask region to assist in feature representation learning. To refine the segmentation mask during testing, we also design a non-parameter adaptive point selection module (APS) to select representative point prompts from query predictions and feed them back to SAM to refine inaccurate segmentation results. Experiments on four standard CD-FSS datasets demonstrate that our method establishes new state-of-the-art results. Code: https://github.com/CVL-hub/GPRN.
\end{abstract}

%

\section{Introduction}
\label{sec1}
With the development of deep convolutional neural networks trained on large-scale datasets~\cite{cordts2016cityscapes, zhou2017scene}, numerous tasks~\cite{li2018occlusion, li2020learning,sun2022coarse} have made promising progress. However, data annotation is a resource-intensive and time-consuming process, particularly when it involves dense pixel-wise labeling for segmentation tasks~\cite{sun2022coarse}. Moreover, models trained on one category typically struggle to generalize to unseen categories, making it even more difficult to scale annotation efforts for diverse classes. In light of this, researchers propose few-shot segmentation (FSS), where the model can adapt to new tasks using a small number of samples \cite{xie2021scale,peng2024multi}.\par 

Existing FSS methods can be broadly classified into two categories: prototype-based~\cite{wang2019panet, liu2022dynamic, yang2020prototype} and correlation-based~\cite{peng2023hierarchical, hong2022cost, moon2023msi}. Prototype-based methods extract prototypes from the support set and generate predictions by comparing the query features with these prototypes~\cite{tian2020prior, yang2020prototype}. Correlation-based methods model the pixel-to-pixel relationship between the query and support features to activate the query pixels that are similar to the support foreground pixels \cite{min2021hypercorrelation, shi2022dense}. However, these methods are limited in scenarios where the base and novel classes are within the same domain. 

\begin{figure}[!t]
\centering
\includegraphics[width=1.0\linewidth]{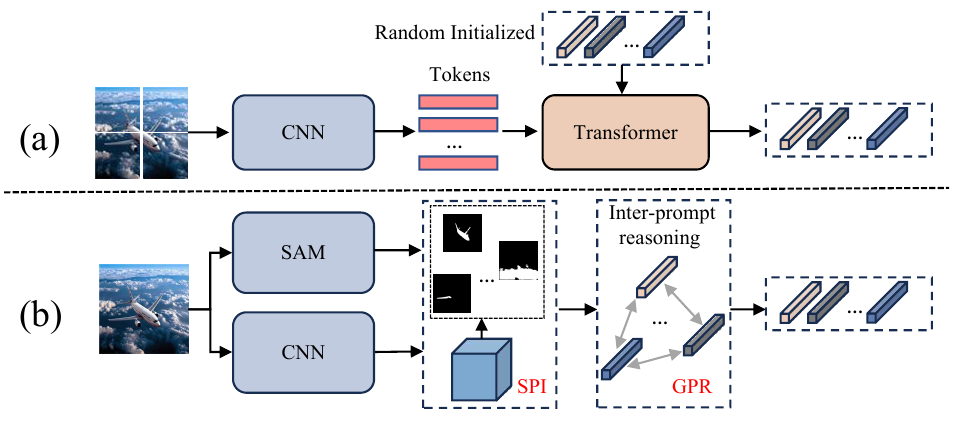}
\caption{\textbf{Comparison of existing visual prompting methods and ours.} (a) Existing methods typically input randomly initialized visual prompts into the network alongside image tokens, which lack prior semantic information and spatial information. (b) Our approach leverages SAM to initialize task-specific visual prompts and constructs a graph convolutional network (GCN) to reason about their inherent relationships. Zoom in for details.}
\vspace{-0.3cm}
\label{motivation}
\end{figure}

To overcome the limitations of traditional FSS, Lei et al. \shortcite{lei2022cross} extend the FSS task to cross-domain few-shot segmentation (CD-FSS) task, where the training and testing datasets belong to different domains. Under this setting, many subsequent works \cite{su2024domain, huang2023restnet} design sophisticated training strategies to learn generalized feature representations from limited data. However, these efforts are constrained by the feature representation capabilities of backbone networks~\cite{he2024apseg} pre-trained on ImageNet~\cite{deng2009imagenet}. To address these limitations, visual prompt tuning (VPT)~\cite{jia2022visual} emerges as a promising solution. VPT introduces learnable prompts into the input space, guiding the model to focus on the most relevant features for the specific task. However, as in Fig.~\ref{motivation} (a), existing VPT methods still face several challenges in CD-FSS. First, they lack prior semantic information during initialization, without any knowledge of the task-specific semantic structures, which can lead to sub-optimal feature adaptation. Moreover, these prompts lack spatial information \cite{wu2022unleashing} and are restricted to transformer architectures. Given that CD-FSS is a dense and semantically-aware task, directly applying this form of visual prompts is sub-optimal.

To address the aforementioned challenges, we resort to the large vision model, i.e., SAM~\cite{kirillov2023segment}, for tackling CD-FSS, which has intrinsic cross-domain generalization ability that provides rich prior knowledge for low-data regimes~\cite{liu2023matcher, zhang2023personalize}. Specifically, we propose a \textbf{G}raph \textbf{P}rompt \textbf{R}easoning \textbf{N}etwork (\textbf{GPRN}) which incorporates SAM to generate \textit{semantic-aware} and \textit{spatial-aware} learnable visual prompts to assist effective feature representation learning for CD-FSS. Our visual prompting scheme is shown in Fig.~\ref{motivation} (b), we first transform the masks generated by SAM into visual prompts enriched with prior knowledge. These masks, which segment objects and capture specific semantic information, enrich each prompt with high-level semantic details for the current task. This leads to our \textbf{S}AM-aware \textbf{P}rompt \textbf{I}nitialization module (\textbf{SPI}). However, SAM prefers to divide an object into many sub-regions, as seen in the aircraft's wing and fuselage in Fig.~\ref{motivation} (b). This may lead to two issues: i) prompts belonging to the same category could be inconsistent, and ii) prompts from small masks contain fragmented information. As such, we propose a \textbf{G}raph \textbf{P}rompt \textbf{R}easoning module (\textbf{GPR}) to mine inter-prompt relationships by constructing a graph among these prompts, enabling prompt interactions and ensuring global semantic consistency among prompts. Subsequently, each visual prompt embeds its semantic information into the corresponding mask region to assist in feature representation learning by the proposed reverse masked average pooling operation. Additionally, to address incomplete segmentation and background interference, we introduce an \textbf{A}daptive \textbf{P}oint \textbf{S}election module (\textbf{APS}) that selects positive and negative point prompts from the initial predictions and feeds them back to the SAM for a final prediction. To sum up, our contributions are as follows:
\begin{itemize}
   \item We propose a SAM-aware graph prompt reasoning network (GPRN), which incorporates SAM to learn visual prompts that integrate both semantic and spatial information to tackle CD-FSS. We further leverage a graph convolutional network (GCN) to reason about the inter-prompt relationships in pursuit of global semantic consistency. To the best of our knowledge, this is the first work to do this in CD-FSS.
   \item We propose an APS module that selects positive and negative prompts from initial predictions to address incomplete foreground segmentation and background interference, thereby improving segmentation accuracy.
   \item Our method achieves state-of-the-art performance on four standard CD-FSS benchmarks.
\end{itemize}

\section{Related Work}
\textbf{Cross-Domain Few-Shot Segmentation.} The CD-FSS setting entails the distribution shifts of data between the base classes during the training phase and the novel classes during the testing phase. Lei et al. \shortcite{lei2022cross} establish a widely used benchmark and design a pyramid transformation module to map features into a domain-invariant space, thus enabling the effective tackling of distribution shifts. Huang et al. \shortcite{huang2023restnet} build on their previous work and observe that the intra-domain relationship between the transformed support and query features can be preserved through the use of residual connections. They resort to learning generalized feature representations from limited source data, but they are constrained by the weak feature representation capabilities of the backbone network. Notably, He et al. \shortcite{he2024apseg} develop an end-to-end segmentation framework based on SAM for CD-FSS, but they directly use SAM as a strong feature extractor and focus on learning high-quality prompts for SAM to avoid the instability associated with manual prompts. In contrast, GPRN incorporates SAM to generate visual prompts that assist with feature representation learning and refine coarse predictions under its guidance.

\noindent \textbf{Prompt Learning.} Prompt learning is first proposed in the field of natural language processing (NLP) as a strategy to adapt pre-trained language models to downstream tasks \cite{brown2020language, lee2018pre}. This advanced paradigm is also introduced into visual tasks \cite{du2022learning, sandler2022fine}. Since manual prompt design requires specialized knowledge, recent works \cite{jia2022visual, wu2022unleashing} resort to optimizing prompts through gradient back-propagation, known as visual prompt tuning. This approach witnesses tremendous success in few-shot scenarios \cite{chen2023semantic, hossain2024visual}, as it allows the model to adapt to new tasks. However, compared to textual prompts in NLP, visual prompts inherently lack high-level semantic information. Additionally, most visual prompt tuning methods are restricted to transformer architectures~\cite{jia2022visual, hossain2024visual}, where visual prompts interact with image tokens to capture their long-range spatial dependencies. To address these challenges, we incorporate SAM to design semantic-aware and spatial-aware visual prompts, aiming to effectively guide the model for CD-FSS tasks.

\noindent \textbf{Graph Convolutional Networks.} Graph convolutional networks (GCNs) excel at handling irregular graph-structured data. They represent each input data point as a node and the adjacency relationships between data points as edges. Through information propagation, each node aggregates the features of its neighboring nodes. This enables GCNs to extract richer and more comprehensive contextual information. GCNs have applications in tasks such as video understanding \cite{wang2018videos, wang2019zero} and zero-shot learning \cite{xie2019attentive, xie2020region}. In this work, we use a variant of GCN called graph attention network (GAT) ~\cite{velickovic2017graph}. The key difference between GAT and GCN is that in GCN, the weights of the relationships between neighboring nodes are precomputed, whereas GAT dynamically assigns weights to each neighboring node. This dynamic weighting allows GAT to better handle complex relationships between nodes. In this way, the inter-prompt relationships can be fully captured to produce high-quality visual prompts for CD-FSS tasks. 

\section{SAM-Aware Graph Prompt Reasoning Network}
\subsection{Task Definition}
The task definition for CD-FSS~\cite{lei2022cross} is summarized as follows. We have a source domain \((X_s, Y_s)\) and a target domain \((X_t, Y_t)\), where $X_*$ represents the data distribution and $Y_*$ denotes the label space. Both the images and the label spaces are disjoint and satisfy $X_s \neq X_t$, $Y_s \cap Y_t = \emptyset$. In this setup, we construct the training set $D_{train}$ from $(X_s, Y_s)$ and test set $D_{test}$ from $(X_t, Y_t)$. Each set consists of multiple episodes. Each episode includes a support set $S=\{(I_k^s, M_k^s)\}_{k=1}^{K}$ and a query set $Q=\{(I^q, M^q)\}$, where $I^*$ and $M^*$ represent the input image and its corresponding ground-truth mask, respectively. Here, $K$ represents the number of support samples. During the training phase, the model is trained on $D_{train}$ to update its parameters. During the test phase, the model's parameters are frozen to evaluate its performance on $D_{test}$.

\subsection{Method Overview}
\begin{figure*}[!t]
\centering
\includegraphics[width=\linewidth]{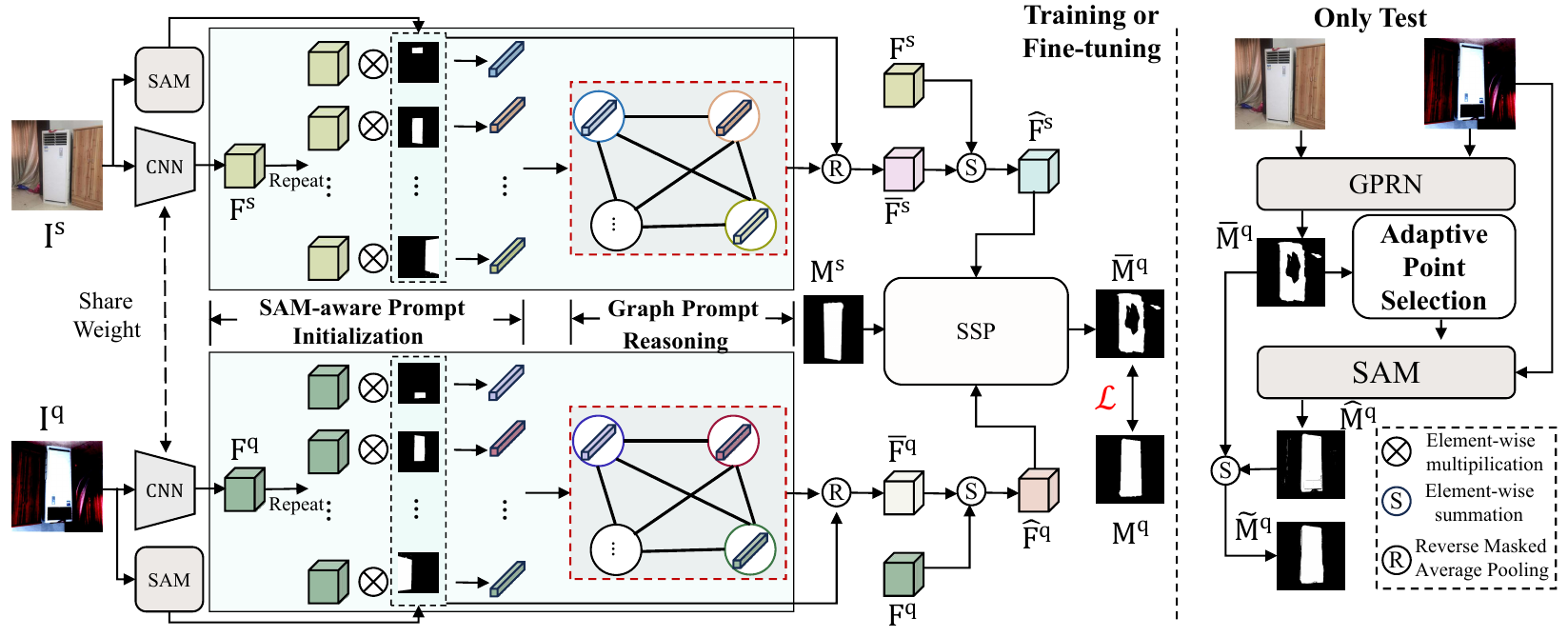}
\caption{\textbf{Overall architecture of our method.} In the training or fine-tuning phase, support and query features along with their corresponding masks generated by SAM are first fed into the SAM-aware initialization module to create visual prompts. These prompts are then processed through a constructed graph to reason about their inter-prompt relationships. Finally, SSP \cite{fan2022self} is employed to segment the query image. In the testing phase, the proposed adaptive point selection module allows for the generation of more accurate segmentation results.}
\vspace{-0.3cm}
\label{framework}
\end{figure*}

Our proposed GPRN is illustrated in Fig. \ref{framework}. The overall architecture of GPRN consists of three main modules: SAM-aware prompt initialization (SPI), graph prompt reasoning (GPR), and adaptive point selection (APS). SPI and GPR are plug-and-play and can be used in the training and/or fine-tuning phase, while APS operates exclusively during testing. Since the operations performed on the support and query in the SPI and GPR are identical, for simplicity, we use $F$ ($\bar{F}$, $\hat{F}$) to represent the feature maps, omitting the superscripts ($s$ or $q$), and denote the masks generated by SAM as $\{m_i\}_{i=1}^{l}$, where $l$ is the number of the masks.

SPI is responsible for generating task-specific visual prompts. It takes as input the features $F$ extracted by CNN and masks $\{m_i\}_{i=1}^{l}$ generated by SAM, and then produces initialized visual prompts, each of which represents the semantic information of its corresponding masked region. Since SAM tends to segment a single object into multiple sub-regions, there may be inconsistencies in semantics between these sub-regions. To address this, GPR constructs a fully connected graph among the visual prompts, where each visual prompt is treated as a node. Through message passing within the graph, each visual prompt aggregates the semantic information of similar nodes, thereby achieving global semantic consistency. Then, a reverse masked average pooling (RMAP) operation is employed to restore the spatial information of the visual prompts. This makes the visual prompts have the same spatial resolution as the feature maps, enabling easy interaction with features. Next, the fused feature map $\hat{F}$ along with support mask $M^s$ are fed into SSP \cite{fan2022self} for prototype-based prediction, resulting in initial query prediction $\bar{M}^q$. The APS is designed to further enhance the final segmentation results. It selects representative geometry points from the initial prediction’s foreground and background regions as point prompts for SAM. SAM then generates predictions specific to these point prompts. Finally, we fuse the initial segmentation result $\bar{M}^q$ with SAM's prediction $\hat{M}^q$ to obtain the final prediction $\Tilde{M}^q$. In the following, we specifically explain the proposed modules: SPI, GPR, and APS.

\subsection{SAM-aware Prompt Initialization}
\label{sec:promptinit}
In previous visual prompt tuning methods~\cite{jia2022visual, wu2022unleashing}, visual prompts are typically initialized randomly and do not contain prior information, making it difficult for them to adapt to the current task. To address this issue, we convert object masks produced by SAM into visual prompts with prior semantic information. However, although the masks $\{m_i\}_{i=1}^{l}$ can cover different semantic objects, there are inevitably overlapping areas among them. We need to eliminate these overlapping regions to ensure that the information within each mask remains consistent and unconfused~\cite{li2024mask}. To resolve this issue, we first sort the masks by area, assigning the smallest index to the mask with the largest area and ensuring that each mask has a unique index. Then, for any overlapping region, we identify the mask with the largest index that includes this region, denoted as $j$. The overlapping part can be eliminated by:
\begin{equation}
\label{overlap}
m_{i}(x,y) =
\begin{cases} 
1, & \text{if} \ i = j \\
0, & \text{if} \ i < j\\
\end{cases}
,
\end{equation}
where $(x,y)$ denotes the spatial coordinates of overlapping region. Eq. (\ref{overlap}) ensures that the overlapping region is assigned to the mask with the largest index (i.e., smallest area), as it has finest information granularity. Meanwhile, it also removes the overlapping region from the larger mask. Next, we use the masked average pooling (MAP) operation to generate the initial visual prompt $v_i$ w.r.t. $m_i$:
\begin{equation}
\label{MAP}
v_i = \text{MAP}(F, m_i) \in \mathbb{R}^{1 \times c},
\end{equation}
where $c$ is the channel dimension. Visual prompts $\{v_i\}_{i=1}^{l}$ are obtained by repeatedly applying Eq. (\ref{MAP}) to $\{m_i\}_{i=1}^{l}$.

\subsection{Graph Prompt Reasoning}
\label{sec:reasoning}
Since SAM tends to over-segment images, i.e., dividing a single object into many sub-regions. This leads prompts belonging to the same category to be inconsistent and prompts from small masks to contain fragmented information. Therefore, it is crucial to explore relationships between visual prompts to allow each prompt to aggregate information from other similar prompts. We use a graph attention network (GAT) to model relationships between visual prompts, where each visual prompt is represented as a node in the graph. GAT's flexible weighting mechanism allows for dynamic adjustment of the relationships between nodes. We first apply learnable matrix $W \in \mathbb{R}^{c \times c}$ to each node to obtain new feature representation, i.e., $\{Wv_i\}_{i=1}^{l}$. Next, we calculate the importance of node $j$ w.r.t. node $i$:
\begin{equation}
\label{edge}
\phi_{ij} = \frac{d(Wv_i, Wv_j)}{\sum_{j=1}^l d(Wv_i, Wv_j)},
\end{equation}
where $d(\cdot, \cdot)$ is the cosine similarity function and $\phi_{ij}$ is the edge weight between node $i$ and $j$, with larger values indicating a stronger correlation between the two nodes. Next, GAT aggregates the weighted average of all neighboring nodes of node $i$ and applies a residual connection to form the new feature representation of node $i$:
\begin{equation}
\label{agg}
\bar{v}_i = v_i + \alpha \sum_{j=1}^{l} \phi_{ij} Wv_j,
\end{equation}
where $\alpha$ is a scaling parameter. By repeatedly applying Eq. (\ref{agg}), we obtained the desirable visual prompts $\{\bar{v}_i\}_{i=1}^{l}$ with global semantic consistency.

\subsection{Self-support Prototype Matching}
Inspired by ~\cite{li2024mask}, we first utilize the reverse process of MAP, referred to as RMAP, to restore the spatial information of visual prompts. This allows us to overcome the limitations of the transformer architecture, enabling visual prompts to interact with features without relying on the attention mechanism:
\begin{equation}
\label{rMAP}
\bar{f}_i = \text{RMAP}(\bar{v}_i,m_i) = \frac{\sum_{x,y}^{h,w} \bar{v}_i^{\mathsf{T}}\otimes m_i(x,y)}{\sum_{x,y}^{h,w} m_i(x,y)},
\end{equation}
where $\otimes$ denotes element-wise multiplication, $\bar{v}_i^{\mathsf{T}}$ is reshaped by $\bar{v}_i$, $\bar{f}_i \in \mathbb{R}^{c \times h \times w}$ represents the feature map w.r.t. the $i$-th visual prompt. Since Eq. (\ref{overlap}) has eliminated the overlapping regions, the resulting feature map $\bar{F} = \sum_{i}^{l} \bar{f}_i$ aggregates clear information from all visual prompts. Finally, we directly summarize $\bar{F}$ and $F$, and the resulting new feature map $\hat{F}$ is used to adapt to the new task. Since the feature representation capability of the CNN is insufficient, $\bar{F}$ provides task-specific feature adaptation. Next, for clarity, we reuse $\hat{F}^s$ and $\hat{F}^q$ to represent new support and query features. SSP \cite{fan2022self} is a prototype-based method. Its main process involves using support prototypes to estimate the query mask and then using query prototypes to predict query features. Since features of the same object are more similar than those between different objects, it helps mitigate the intra-class variation problem. The above process can be formulated as:
\begin{align}
\label{ssp}
P^s &= \text{MAP}(\hat{F}^s, M^s)\in \mathbb{R}^{2 \times c}, \\
\bar{M}^q & = \text{SSP}(\hat{F^q}, P^s) \in \mathbb{R}^{2 \times h \times w},
\end{align}
where $P^s$ is the support prototypes, $\bar{M}^q$ is the query prediction, $h$ and $w$ represent height and width, respectively.

\subsection{Adaptive Point Selection}
\label{sec:MaskSLIC}
After obtaining the initial prediction $\bar{M}^q$, its accuracy is often suboptimal. Therefore, we aim to refine $\bar{M}^q$ using SAM. Our method involves selecting positive and negative point prompts from both foreground and background mask regions as inputs for the SAM prompt mode. This directs SAM to enhance the segmentation of partially segmented target regions and reduce background interference. Since the process for selecting prompts from both the foreground and background are identical, we only describe the procedure for choosing positive prompts.

To make the prompts more representative w.r.t. mask regions (i.e., to cover the broadest possible area with the fewest numbers), inspired by maskSLIC \cite{irving2016maskslic}, we start by using a bounding box to enclose the irregular mask, resulting in region $R$. We denote the set of points within the mask as $E$ and those outside the mask as $O$, where $R = E \cup O$, and initialize $ Z = \{\}$ to store the selected points. For each $e \in E$, its distance transform $DT(e)$ can be formulated as:
\begin{equation}
\label{dtrans}
DT(e) = \min_{o \in O} \left( \sum_{i=1}^{n} (e_i - o_i)^2 \right)^{\frac{1}{2}},
\end{equation}
where, $n$ denotes the feature dimension, and in this case, $n = 2$ because $e$ and $o$ represent two-dimensional coordinates. Next, we apply Gaussian smoothing to $DT(E)$ to reduce the impact of edge noise:
\begin{equation}
\label{gussian}
DT(E) = \text{Gaussian}(DT(E), \gamma),
\end{equation}
where $\gamma$ is a scalar parameter that controls the extent of smoothing, and we set it to 0.1 according to \cite{irving2016maskslic}. The point $e$ that maximizes $DT(e)$ is the current candidate point we have found:
\begin{equation}
\label{candidate}
e_* = \arg \max_e DT(e).
\end{equation}
We remove $e_*$ from $E$, $O$ becomes $O \cup \{e_*\}$, and $Z$ becomes $Z \cup \{e_*\}$ for the next iteration. At the end of the iteration, we obtain the desirable positive point prompts $Z$. Points located at the edges of the mask exhibit significant instability. Eq. (\ref{gussian}) assigns greater weights to points situated near the center, thereby ensuring that the $e_*$ is positioned away from the edges. Simultaneously, Eq.(\ref{candidate}) ensures that the candidate points in $Z$ are as spatially distant from each other as possible, thereby maximizing the coverage area of each point. Negative prompts $N$ can be derived from the background mask in $\bar{M}^q$ using Eq. (\ref{dtrans}) to (\ref{candidate}).

Subsequently, we input $Z$ and $N$ into the SAM along with the query image $I^q$. In prompt mode, the SAM outputs the segmentation logits $\hat{M}^q_{lgt} \in \mathbb{R}^{1 \times h \times w}$.
The predicted probability map $\hat{M}^q$ is obtained by using the following formulas:
\begin{equation}
\label{promap}
\hat{M}^q = (\mathbf{1} - \sigma(\hat{M}^q_{lgt})) \oplus \sigma (\hat{M}^q_{lgt}) \in \mathbb{R}^{2 \times h \times w}.
\end{equation}
Here, $\sigma$ represents the sigmoid function, and $\mathbf{1}$ denotes a matrix of ones with the same shape as $\hat{M}^q_{lgt}$. The symbol $\oplus$ indicates concatenation along the channel dimension. Finally, we fuse $\bar{M}^q$ with $\hat{M}^q$ to obtain the final result $\Tilde{M}^q$:
\begin{equation}
\label{refine}
\Tilde{M}^q = \beta \hat{M}^q + (1 - \beta)\bar{M}^q,
\end{equation}
where $\beta$ is a scaling parameter.
\subsection{Loss Function}
Since the APS module is used only during testing, the final prediction of the model during training and/or fine-tuning is $\bar{M}^q$. Using the binary cross entropy (BCE) function, the model parameters are updated by optimizing the following loss:
\begin{equation}
\label{loss}
\mathcal{L} = \text{BCE}(\bar{M}^q, M^q).
\end{equation}

\section{Experiments}

\subsection{Datasets and Implementation Details}
Our model can be optionally trained on the base data, i.e., PASCAL VOC \cite{everingham2010pascal} augmented with SDS \cite{hariharan2011semantic}. We fine-tune and test it on four datasets: Deepglobe \cite{demir2018challenge}, ISIC \cite{codella2019skin}, Chest X-Ray \cite{candemir2013lung}, and FSS-1000 \cite{li2020fss}. 
We use mean-IoU as the performance evaluation metric. It is calculated by averaging the intersection-over-unions (IoUs) across different foreground classes. 

\begin{table*}[t]
\setlength{\tabcolsep}{4pt}
 \renewcommand\arraystretch{1.2}
    \centering
    \begin{tabular}{c|c|c|cc|cc|cc|cc|cc}
        \toprule[1pt]
        \multirow{2}*{Methods} & \multirow{2}*{Backbone} & \multirow{2}*{Publication} & \multicolumn{2}{c|}{Deepglobe} & \multicolumn{2}{c|}{ISIC} & \multicolumn{2}{c|}{Chest X-Ray} & \multicolumn{2}{c|}{FSS-1000} & \multicolumn{2}{c}{\textbf{Average}} \\
        & & & 1-shot & 5-shot & 1-shot & 5-shot & 1-shot & 5-shot & 1-shot & 5-shot & 1-shot & 5-shot \\
        \midrule
        \multicolumn{13}{c}{Methods without Fine-tuning Phase} \\
        \midrule
         AMP \shortcite{siam2019amp} & Vgg-16 & ICCV & 37.6 & 40.6 & 28.4 & 30.4 & 51.2 & 53.0 & 57.2 & 59.2 & 43.6 & 45.8 \\
          RestNet \shortcite{huang2023restnet} & ResNet-50 & BMVC& - & - & 42.3 & 51.1 & 70.4 & 73.7 & \underline{81.5} & \underline{84.9} & - & - \\
        PerSAM \shortcite{zhang2023personalize} & ViT-base & ICLR& 30.0 & 30.1 & 23.3 & 25.3 & 61.0 & 66.5 & 36.1 & 40.7 & 37.6 & 40.6 \\
        ABCDFSS \shortcite{herzog2024adapt} & ResNet-50 & CVPR & 42.6 & 49.0 & 45.7 & 53.3 & 79.8 & 81.4 & 74.6 & 76.2 & 60.7 & 65.0 \\
        PMNet \shortcite{chen2024pixel} & ResNet-50 & WACV& 37.1 & 41.6 & 51.2 & 54.5 & 70.4 & 74.0 & \textbf{84.6} & \textbf{86.3} & 60.8 & 64.1 \\
        DRANet \shortcite{su2024domain} & ResNet-50 & CVPR& 41.3 & 50.1 & 40.8 & 48.8 & 82.4 & 82.3 & 79.1 & 80.4 & 60.9 & 65.4 \\
        APSeg \shortcite{he2024apseg} & ViT-base & CVPR& 36.0 & 40.0 & 45.4 & 54.0 & \underline{84.1} & 84.5 & 79.7 & 81.9 & 61.3 & 65.1 \\
        \midrule
        \multicolumn{13}{c}{Methods with Fine-tuning Phase} \\
        \midrule
        PATNet \shortcite{lei2022cross} & ResNet-50 &ECCV & 37.9 & 43.0 & 41.2 & 53.6 & 66.6 & 70.2 & 78.6 & 81.2 & 56.1 & 62.0 \\
        DARNet \shortcite{fan2023darnet} & ResNet-50 & Arxiv& 44.6 & 54.1 & 47.8 & 60.5 & 81.2 & \textbf{89.7} & 76.4 & 83.2 & 62.5 & \underline{71.9} \\
        IFA \shortcite{nie2024cross} & ResNet-50 & CVPR& \underline{50.6} & \underline{58.8} & \underline{66.3} & \underline{69.8} & 74.0 & 74.6 & 80.1 & 82.4 & \underline{67.8} & 71.4 \\
        DMTNet \shortcite{ijcai2024p71} & ResNet-50 & IJCAI& 40.1 & 51.2 & 43.6 & 52.3 & 73.8 & 77.0 & 81.5 & 83.3 & 59.7 & 66.0 \\
        GPRN (ours) & ResNet-50 &- & \textbf{51.7} & \textbf{59.3} & \textbf{66.8} & \textbf{72.2} & \textbf{87.0} & \underline{87.1} & 81.1 & 82.6 & \textbf{71.7} & \textbf{75.3} \\
        \bottomrule[1pt]
    \end{tabular}
    \caption{Comparison of model performance in 1-shot and 5-shot settings with and without fine-tuning. The best and second-best methods are highlighted in \textbf{bold} and \underline{underlined}, respectively.}
    \label{tab:performance}
    \vspace{-0.4cm}
\end{table*}
\begin{figure*}[!t]
\centering
\includegraphics[width=1.0\linewidth]{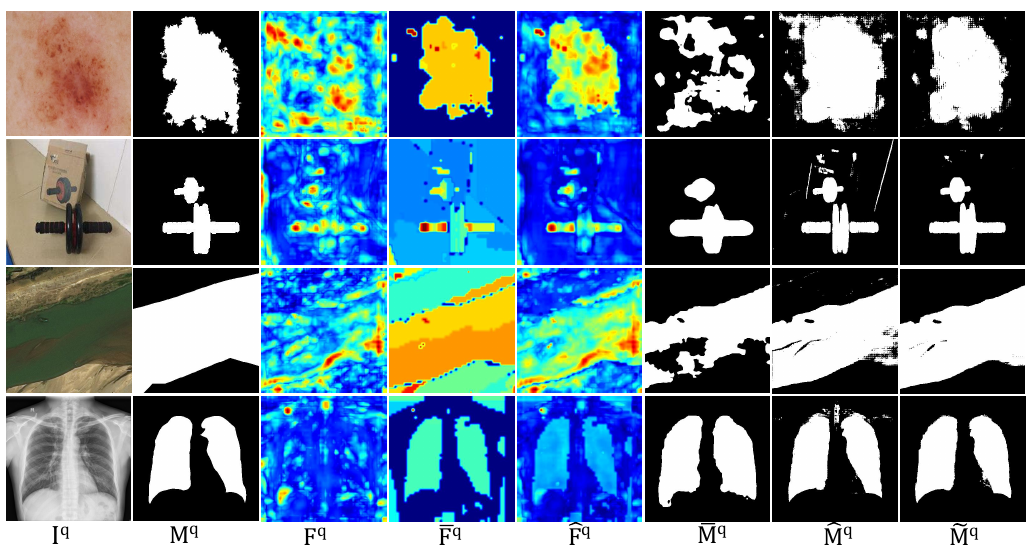}
\caption{Qualitative analysis results: $I^q$ and $M^q$ represent the original query image and its ground truth mask, respectively. $F^q$, $\bar{F}^q$, $\hat{F}^q$ correspond to the feature map extracted by the backbone network, the feature map of the visual prompts, and the feature map adapted to the new task, respectively. $\bar{M}^q$, $\hat{M}^q$, $\Tilde{M}^q$ represent the model's prediction, SAM's prediction, and the final segmentation result after refinement, respectively.}
\label{vis}
\vspace{-0.5cm}
\end{figure*}
Our baseline model is SSP \cite{fan2022self}, utilizing the ResNet-50 backbone pre-trained on ImageNet \cite{deng2009imagenet}. To minimize computational overhead, we employ the base version of SAM. All images are initially resized to 400$\times$400 for input into the CNN and subsequently upsampled to 1024$\times$1024 to meet SAM’s input requirements. The number of masks $l$ is fixed as 40. If SAM generates more than 40 masks, we discard those with the smallest areas; if fewer than 40 masks are generated, we pad them with zero masks. The parameter $\alpha$ is set to 0.1. The numbers of positive and negative point prompts, $\left| Z \right|$ and $\left| N \right|$, are both set to 20, and $\beta$ is set to 0.5. During training, our model is optimized with 0.9 momentum and an initial learning rate of 1e-3 and training with 5 epochs. 
During fine-tuning, we use 120, 520, 60, and 20 samples from the Deepglobe, FSS-1000, ISIC, and Chest X-ray datasets, respectively. We employed the SGD optimizer with a learning rate of 5e-4, momentum of 0.9, and weight decay of 5e-4. For evaluation, we randomly sample 600 episodes from ISIC, Deepglobe, and Chest X-Ray, and 2400 episodes from FSS-1000, averaging the results over five random seeds.

\subsection{Comparisons with State-of-the-arts}
Tab. \ref{tab:performance} presents a comparison of GPRN with previous state-of-the-art CD-FSS methods under 1-shot and 5-shot settings. To ensure a fair comparison, we categorize the methods based on whether they include a fine-tuning phase, as fine-tuning typically enhances performance. GPRN demonstrates a significant advantage, outperforming the best non-fine-tuning method by 10.4\% in the 1-shot setting and 9.9\% in the 5-shot setting in terms of the mean-IoU. Even compared to methods that include a fine-tuning phase, GPRN surpasses the top-performing fine-tuning method (IFA) by 3.9\% in both 1-shot and 5-shot settings. This establishes GPRN as the new state-of-the-art in various CD-FSS benchmarks. Additionally, when compared to other SAM-based methods such as PerSAM \cite{zhang2023personalize} and APSeg \cite{he2024apseg} for the 1-shot setting, GPRN achieves significant improvements of 34.1\% and 10.4\%, respectively. These results highlight the effectiveness of the proposed modules in fully leveraging SAM’s potential in CD-FSS.

Our model can be flexibly extended to the $K$-shot scenario by averaging the prototypes from the support set, while all subsequent steps remain consistent with the $1$-shot setting. We find that GPRN achieves good results even when fine-tuning using only a small amount of data, so we report only the fine-tuning results in Tab. \ref{tab:performance}.

\begin{table*}[htbp]
\centering
\begin{minipage}{0.32\textwidth}
\centering
\setlength{\abovecaptionskip}{5pt}
\setlength{\belowcaptionskip}{5pt}
\renewcommand\arraystretch{1.2}
\resizebox{0.8\linewidth}{!}{
\begin{tabular}{ccc|c} 
\toprule
SPI & GPR & APS & 1-shot mean-IoU \\
\midrule
         &     &  & 73.8  \\
$\checkmark$   &     & & 77.9      \\
 &     &  $\checkmark$ & 77.2  \\
   $\checkmark$     & $\checkmark$   &  & 78.4 \\
   $\checkmark$      &     &  $\checkmark$ & 80.7  \\
$\checkmark$   & $\checkmark$   & $\checkmark$ & 81.1 \\ 
\bottomrule
\end{tabular}}
\caption{Effects of different modules.}
\label{component}
\end{minipage}
\hfill
\begin{minipage}{0.32\textwidth}
\centering
\setlength{\abovecaptionskip}{5pt}
\setlength{\belowcaptionskip}{5pt}
\renewcommand\arraystretch{1.2}
\centering
\resizebox{0.8\linewidth}{!}{
\begin{tabular}{c|c} 
\toprule
$\alpha$ & 1-shot mean-IoU \\
\midrule
   0.1      &   81.1  \\
   0.3      &    80.9 \\
   0.5      &    80.5 \\
\bottomrule
\end{tabular}}
\caption{Effects of different $\alpha$.}\label{alpha}
\end{minipage}
\hfill
\begin{minipage}{0.32\textwidth}
\centering
\setlength{\abovecaptionskip}{5pt}
\setlength{\belowcaptionskip}{5pt}
\renewcommand\arraystretch{1.2}
\centering
\resizebox{0.8\linewidth}{!}{
\begin{tabular}{c|c} 
\toprule
Number & 1-shot mean-IoU \\
\midrule
   10      &   76.2  \\
   20      &    78.4 \\
   40      &     81.1 \\
   60      &     81.3 \\
\bottomrule
\end{tabular}}
\caption{Effects of number of masks $l$.}\label{masknum}
\end{minipage}
\end{table*}

\begin{table*}[htbp]
\centering
\begin{minipage}{0.32\textwidth}
\centering
\setlength{\abovecaptionskip}{5pt}
\setlength{\belowcaptionskip}{5pt}
\renewcommand\arraystretch{1.2}
\centering
\resizebox{0.8\linewidth}{!}{
\begin{tabular}{cc|c} 
\toprule
$\left| Z \right|$ & $\left| N \right|$& 1-shot mean-IoU \\
\midrule
   10      &   0  &  79.2 \\
   10      &    10 & 80.5 \\
   20      &    20 & 81.1 \\
   30      &    30 & 81.0 \\
\bottomrule
\end{tabular}}
\caption{Effects of number of point prompts $\left| Z \right|$ and $\left| N \right|$
 on FSS-1000.}\label{number}
 \vspace{-0.2cm}
\end{minipage}
\hfill
\begin{minipage}{0.32\textwidth}
\centering
\setlength{\abovecaptionskip}{5pt}
\setlength{\belowcaptionskip}{5pt}
\renewcommand\arraystretch{1.2}
\centering
\resizebox{\linewidth}{!}{
\begin{tabular}{c|c} 
\toprule
Strategy & 1-shot mean-IoU \\
\midrule
   \cite{zhang2023personalize}      & 77.9 \\
   Box      & 75.3 \\
   APS~(ours)       &  81.1 \\
   APS + Box      & 81.1 \\
\bottomrule
\end{tabular}}
\caption{Effects of different prompts selection strategies on FSS-1000.}\label{fine-tuning}
\vspace{-0.2cm}
\end{minipage}
\hfill
\begin{minipage}{0.32\textwidth}
\centering
\setlength{\abovecaptionskip}{5pt}
\setlength{\belowcaptionskip}{5pt}
\renewcommand\arraystretch{1.2}
\centering
\resizebox{0.65\linewidth}{!}{
\begin{tabular}{c|c} 
\toprule
$\beta$ & 1-shot mean-IoU \\
\midrule
   0.1      &    78.5 \\
   0.3      &    80.2 \\
   0.5      &    81.1 \\
   0.6      &    80.2 \\
\bottomrule
\end{tabular}}
\caption{Effects of different $\beta$ on FSS-1000.}\label{beta}
\vspace{-0.2cm}
\end{minipage}
\end{table*}

\subsection{Ablation Study and Visualizations}

\noindent \textbf{Effects of different modules.}
Tab. \ref{component} presents the impact of GPRN's key components on accuracy for the FSS-1000 dataset. The first row represents our baseline. It can be observed that the SPI module provides a significant mean-IoU improvement of 4.1\%. Adding the GPR module further enhances performance, achieving a mean-IoU of 78.4\%. This improvement is attributed to the GPR module's ability to enhance global semantic consistency among prompts. Additionally, the parameter-free APS module contributes a 3.4\% mean-IoU increase. Combining all three modules results in a mean-IoU of 81.1\% (+7.3\%). 
This indicates that the three modules are complementary. It can be interpreted that SPI and GPR enhance the accuracy of the initial predictions, ensuring that APS selects more representative point prompts, thereby further improving segmentation accuracy.

\noindent \textbf{Effects of different $\alpha$.}
We set $\alpha$ to 0.1, 0.3, and 0.5 to investigate its impact on the results. As shown in Tab. \ref{alpha}, the model achieves the best performance when $\alpha$ is 0.1.

\noindent \textbf{Effects of number of masks $l$.}
The number of masks $l$ directly affects the model's performance. When $l$ is small, the masks are coarse-grained, while increasing $l$ results in finer-grained masks. We set $l$ to 10, 20, 40, and 60, with corresponding results shown in Tab. \ref{masknum}. When $l$ = 40 and $l$ = 60, the model's mean-IoU accuracy is similar. Considering both performance and computational efficiency, we set $l$ to 40.

\noindent \textbf{Effects of number of point prompts $\left| Z \right|$ and $\left| N \right|$.}
\label{sec:numberpoint}
The number of point prompts affects both performance and computational efficiency. Comparing the first and second rows in Tab. \ref{number} shows that selecting negative point prompts from the background also improves performance. When the number of point prompts from both the foreground and background is set to 20, the model's performance is saturated. To reduce computational burden, we ultimately select 20 point prompts from both the foreground and background.

\noindent \textbf{Effects of different prompts selection strategies.}
To further validate the necessity of the point prompt selection strategy in APS, in Tab.~\ref{fine-tuning}, we compare the APS with the selection principle proposed in PerSAM \cite{zhang2023personalize}, which selects the points with the highest foreground or background probabilities from the prediction results as candidate point prompts, and the prompt form that directly specifies the target location using a box. We similarly select 20 points for both the foreground and background. Our strategy outperforms the one proposed by PerSAM by a clear margin of 3.2\%. We also observe that using a box along with APS does not bring any performance improvement. So our final prompt solution is to use APS only.

\noindent \textbf{Effects of different $\beta$.}
The selection of the $\beta$ determines the balance between incorporating correction information from SAM and the original prediction data. This is a crucial hyperparameter that can significantly affect the model's performance. We conduct ablation experiments with $\beta$ of 0.1, 0.3, 0.5, and 0.6 in Tab. \ref{beta}. The model achieves the best performance when the $\beta$ value is set to 0.5. As the beta value increases from 0.1 to 0.5, the model's performance improves significantly. However, further increasing the beta value to 0.6 results in a decline in performance. This indicates that there is a complementary relationship between the model's predictions and SAM's predictions, with neither being more accurate than the weighted fusion of the two.

\noindent \textbf{Visualizations}
To better understand the effectiveness of the modules, Fig. \ref{vis} provides some visualizations. It can be found that the feature map $F^q$ extracted by the backbone contains significant aliasing artifacts, while the visual prompts $\bar{F}^q$ exhibit clear boundaries. The combined feature map $\hat{F}^q$ which incorporates information from $\bar{F}^q$ significantly reduces the impact of noise. The initial prediction $\bar{M}^q$ is often inaccurate, and SAM's prediction $\hat{M}^q$ also contains many artifacts. However, $\Tilde{M}^q$, which fuses both predictions, achieves the best segmentation performance.

\section{Conclusion}
In this work, we propose a SAM-aware graph prompt network (GPRN) to tackle the CD-FSS task. Specifically, we transform the masks generated by SAM into several visual prompts, followed by a graph attention network to reason about inter-prompt relationships. We also design a novel parameter-free test-time refine module, which can enhance the model's performance with a clear margin. GPRN achieves state-of-the-art performance on various CD-FSS datasets.


\bibliography{ref}

\begin{strip}
\section*{\textbf{\textit{Supplementary Material} for \\ SAM-Aware Graph Prompt Reasoning Network for Cross-Domain Few-Shot Segmentation}}
\end{strip}
 
In this supplementary material, we detail the implementation of SSP, present comprehensive experiments to validate the effectiveness of our method further, and include additional visualizations to help understand the proposed approach.

\begin{table*}[htb]
    \renewcommand\arraystretch{1.2}
    \centering
    \resizebox{1.0\linewidth}{!}{
    \begin{tabular}{l|c|cc|cc|cc|cc|cc}
        \toprule[1pt]
        \multicolumn{12}{c}{Source Domain: Pascal VOC 2012 $\to$ Target Domain: Below}\\\hline
        \multirow{2}*{Methods}& \multirow{2}*{Backbone}& \multicolumn{2}{c|}{Deepglobe}& \multicolumn{2}{c|}{ISIC}& \multicolumn{2}{c|}{Chest X-Ray}& \multicolumn{2}{c|}{FSS-1000}& \multicolumn{2}{c}{\textbf{Average}}\\\cline{3-12}
        ~& ~& 1-shot& 5-shot& 1-shot& 5-shot& 1-shot& 5-shot& 1-shot& 5-shot& 1-shot& 5-shot\\\hline
        GPRN(w/o fine-tuning)& \multirow{3}*{ResNet-50}& 45.4& 54.0& 53.7& 62.6& 78.7& 79.1& 81.1& 82.5& 64.7& 69.6\\
        GPRN(w fine-tuning)& ~& 49.6& 58.7& 68.9& 75.2& 86.4& 86.9& 82.0& 83.5& 71.7& 76.1\\
        GPRN(final)& ~& 51.7& 59.3& 66.8& 72.2& 87.0& 87.1& 81.1& 82.6& 71.7& 75.3\\
        \bottomrule[1pt]
    \end{tabular}}
    \vspace{-0.3cm}
    \caption{Performance of GPRN trained on the PASCAL VOC with and without the fine-tuning phase, and with only fine-tuning.}
    \label{tab1}
\end{table*}
\section{Implementation Details of SSP}

\begin{figure}[!t]
\centering
\includegraphics[width=1.0\linewidth]{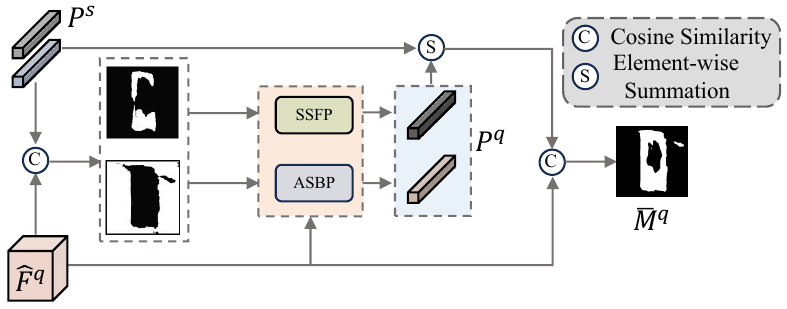}
\caption{Flowchart of SSP. SSFP refers to the self-support foreground prototype while ASBP refers to the adaptive self-support background prototype proposed in SSP, respectively.}
\label{SSP}
\end{figure}

SSP is a prototype-based FSS method, and its core idea is to use query prototypes to match the query feature itself. This approach is inspired by the Gestalt principle, which posits that feature similarity within the same object is higher than that between different objects. It has also been demonstrated that the target datasets in CD-FSS adhere to the Gestalt principle. By aligning query prototypes with the query feature, SSP helps to narrow the feature gap between support and query. Its main process is illustrated in Fig. \ref{SSP}. First, the coarse query prediction is obtained by calculating the cosine similarity between support prototypes and the query feature:
\begin{equation}
\label{coarse}
M_{\text{coarse}} = \text{softmax}(\text{cos}(\hat{F}^q, P^s)) \in  \mathbb{R}^{2 \times h \times w},
\end{equation}
here, $M_{\text{coarse}} = \{M_{\text{coarse}}^b, M_{\text{coarse}}^f\}$, where $ M_{\text{coarse}}^b$ and $M_{\text{coarse}}^f$ denote the coarse probability maps for the background and foreground, respectively. 
Then, SSFP is responsible for generating the query foreground prototype $P^q_{f}$, and its approach is relatively straightforward: a threshold $\tau_{f} = 0.7$ is set to filter out high-confidence foreground regions, and the features within these regions are averaged:
\begin{equation}
\label{SSFP}
P^q_{f} = \text{MAP}(\hat{F}^q, M_{\text{coarse}}^f(x,y) > \tau_{f}) \in \mathbb{R}^{1 \times c}.
\end{equation}
Since the background is more complex, using a single global background prototype to represent it may entangle different semantic concepts. ASBP dynamically aggregates similar background pixels for each query pixel to generate adaptive self-support background prototypes. Specifically, we first gather the background features $\hat{F}^{q,b}$ from high-confidence background regions:
\begin{equation}
\label{ASBP}
\hat{F}^{q,b} = \hat{F}^q \otimes (M_{\text{coarse}}^b(x,y) > \tau_{b}) \in 
 \mathbb{R}^{c \times t}.
\end{equation}
Where $\tau_{b} = 0.6$ is the background threshold, $t$ is the number of pixels within the high-confidence background regions, and $\otimes$ denotes element-wise multiplication.
Then, we calculate the similarity matrix $A$ between $\hat{F}^{q,b}$ and $\hat{F}^{q}$ through matrix multiplication:
\begin{equation}
\label{A}
A = \text{matmul}(\hat{F}^{{q,b}^{\mathsf{T}}}, \hat{F}^{q}) \in \mathbb{R}^{t \times h \times w}. 
\end{equation}
The adaptive self-support background prototypes are derived by:
\begin{equation}
\label{background}
P^q_{b} = \text{matmul}(\hat{F}^{{q,b}}, \text{softmax}(A)) \in \mathbb{R}^{c \times h \times w},
\end{equation}
here, the $\text{softmax}$ operation is performed along the first dimension. To this end, we obtain the desired query prototypes $P^q = \{P^q_{f}, P^q_{b}\}$. We weighted combine the support prototypes $P^s$ and self-support query prototypes $P^q$:
\begin{equation}
\label{weightcom}
P = \alpha1P^s + \alpha2P^q,
\end{equation}
where $\alpha1 = \alpha2 = 0.5$. The final matching prediction $\bar{M}^q$ is generated by computing the cosine distance between the prototypes $P$ and query feature $\hat{F}^q$:
\begin{equation}
\label{final}
\bar{M}^q = \text{softmax}(\text{cos}(\hat{F}^q, P)) \in \mathbb{R}^{2 \times h \times w}.
\end{equation}

\section{Performance of GPRN trained on the base data with and without fine-tuning}
In the main paper, we mention that GPRN is optionally trained on PASCAL VOC and we only report the performance w/o training. Now, we further report the performance of GPRN trained on PASCAL. The result is shown in Tab. \ref{tab1}. We observe that training on PASCAL VOC does not result in any performance improvement, while the fine-tuning phase provides a 4.0\% boost in performance.

\section{Performance of GPRN with and without eliminating the overlapping regions of masks}
The overlapping regions between masks cause information inconsistency, leading to a decline in the quality of visual prompts obtained through RMAP. In Tab. \ref{tab3}, we compare the specific extent of this impact.

\section{More Visualizations}
To better understand the effectiveness of the proposed APS module, Fig. \ref{vis} provides visualizations. In $\bar{M}^q$, the green and red dots represent the positive and negative points selected by APS, respectively. We can observe that their distribution is dispersed and uniform, which prevents the selected geometric points from being overly concentrated and thus avoids information redundancy. Therefore, they can guide SAM to explore potential foreground and background regions, thereby increasing segmentation accuracy.

\begin{figure}[!t]
\centering
\includegraphics[width=1.0\linewidth]{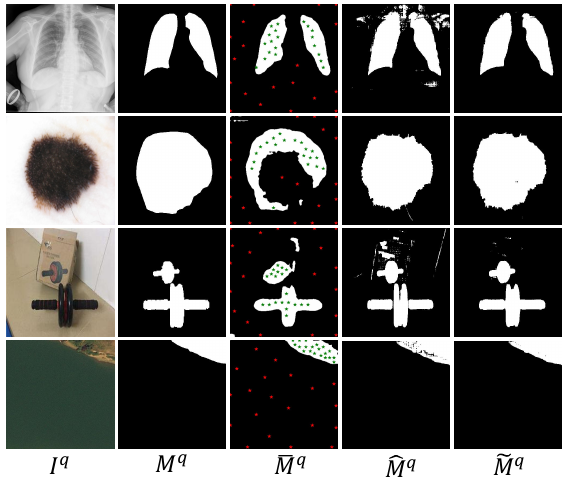}
\vspace{-0.6cm}
\caption{Qualitative analysis results: $I^q$ and $M^q$ represent the original query image and its ground truth mask, respectively.  $\bar{M}^q$, $\hat{M}^q$, $\Tilde{M}^q$ represent the model's prediction, SAM's prediction, and the final segmentation result after refinement, respectively. The green and red dots in $\bar{M}^q$ represent the positive and negative points selected by our APS module, respectively.}
\label{vis}
\end{figure}

\begin{figure}[!t]
\centering
\includegraphics[width=1.0\linewidth]{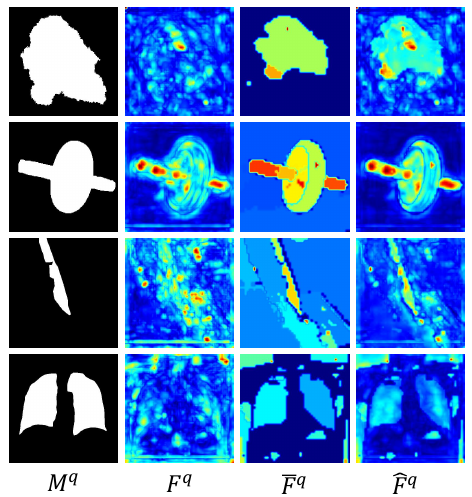}
\vspace{-0.7cm}
\caption{Qualitative analysis results: $M^q$ represent the query ground truth mask. $F^q$, $\bar{F}^q$, $\hat{F}^q$ correspond to the feature map extracted by the backbone network, the 3D feature map of the visual prompts, and the feature map adapted to the new task, respectively.}
\label{vis2}
\end{figure}

Fig. \ref{vis2} provides further evidence that our GPRN enhances feature representation learning. The features $F^q$ extracted by the ImageNet-pre-trained backbone (ResNet-50) are quite disorganized. This is because the backbone struggles to generalize images from different domains and classes that it hasn't encountered during training. In contrast, the large-scale visual model SAM, pre-trained on tens of millions of images from various domains and classes, possesses excellent generalizability. It generates visual prompts ($\bar{F}^q$) that delineate clear boundaries for each semantic object, and the features within each object are highly consistent. This undoubtedly mitigates the limitations in the expressive power of $F^q$. (as seen in $\hat{F}^q$, which combines $F^q$ and $\bar{F}^q$).

\subsubsection{Proposed Module Visualizations}
\begin{figure*}[h]
\centering
\includegraphics[width=1.0\linewidth]{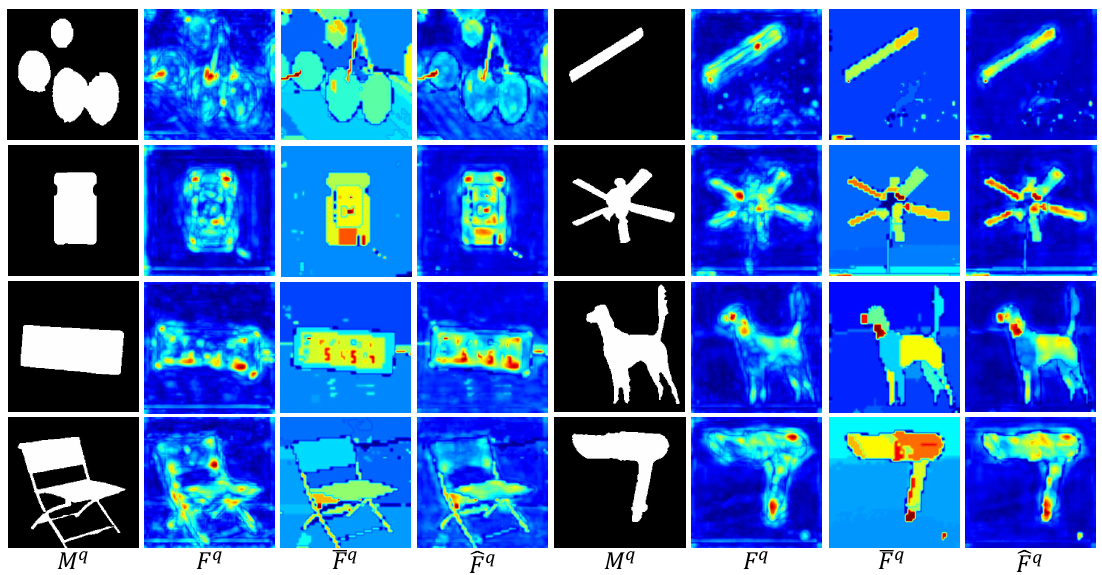}
\caption{Qualitative analysis results on FSS-1000: $M^q$ represent the query ground truth mask. $F^q$, $\bar{F}^q$, $\hat{F}^q$ correspond to the feature map extracted by the backbone network, the 3D feature map of the visual prompts, and the feature map adapted to the new task, respectively.}
\label{vis3}
\end{figure*}

\begin{figure*}[h!]
\centering
\includegraphics[width=1.0\linewidth]{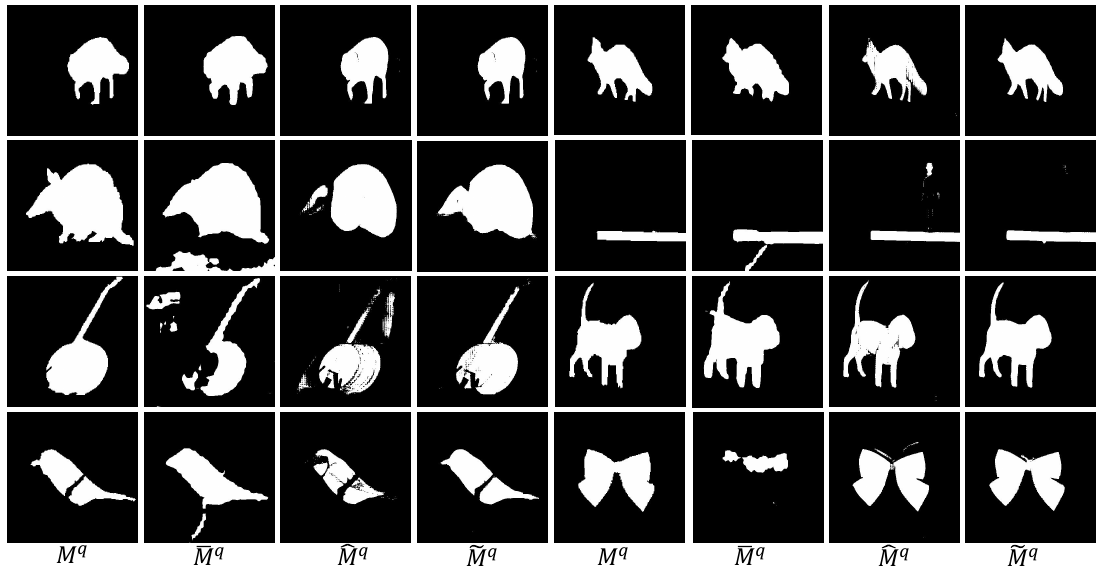}
\caption{Qualitative analysis results on FSS-1000: $M^q$ represent the query ground truth mask. $\bar{M}^q$, $\hat{M}^q$ and $\Tilde{M}^q$ represent the model’s prediction,
SAM’s prediction, and the final segmentation result after refinement, respectively.}
\label{vis4}
\end{figure*}

\begin{figure*}[t!]
\centering
\includegraphics[width=1.0\linewidth]{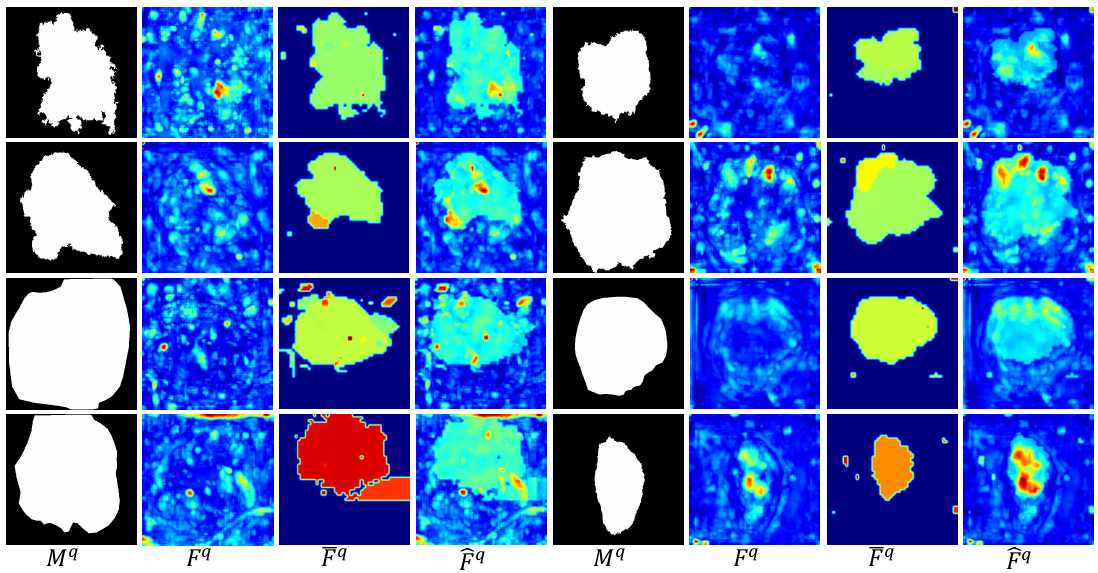}
\caption{Qualitative analysis results on ISIC: $M^q$ represent the query ground truth mask. $F^q$, $\bar{F}^q$, $\hat{F}^q$ correspond to the feature map extracted by the backbone network, the 3D feature map of the visual prompts, and the feature map adapted to the new task, respectively.}
\label{vis5}
\end{figure*}

\begin{figure*}[t!]
\centering
\includegraphics[width=1.0\linewidth]{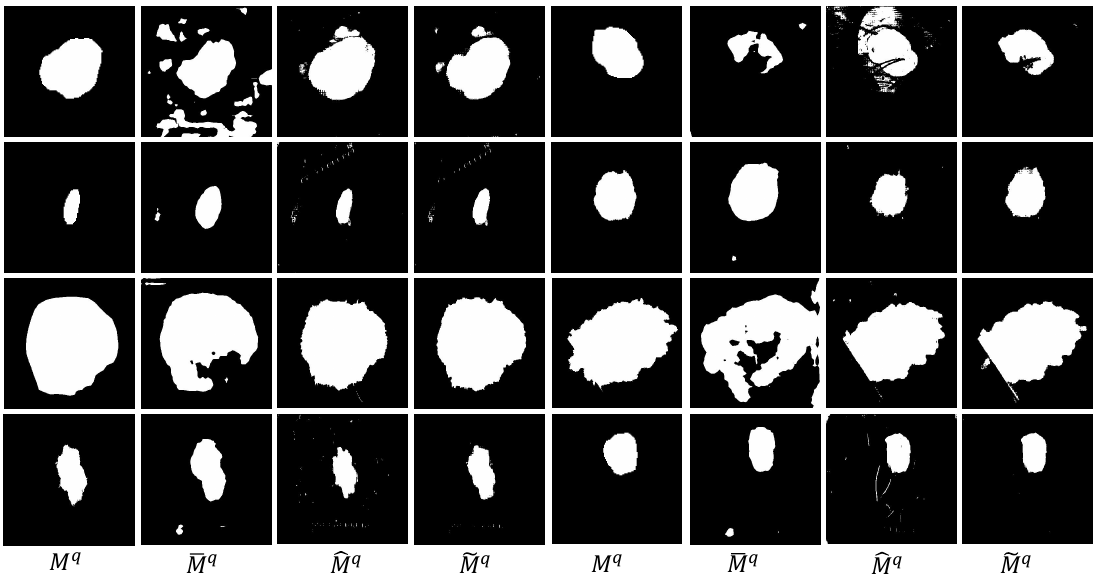}
\caption{Qualitative analysis results on ISIC: $M^q$ represent the query ground truth mask. $\bar{M}^q$, $\hat{M}^q$ and $\Tilde{M}^q$ represent the model’s prediction,
SAM’s prediction, and the final segmentation result after refinement, respectively.}
\label{vis6}
\end{figure*}

\begin{figure*}[t!]
\centering
\includegraphics[width=1.0\linewidth]{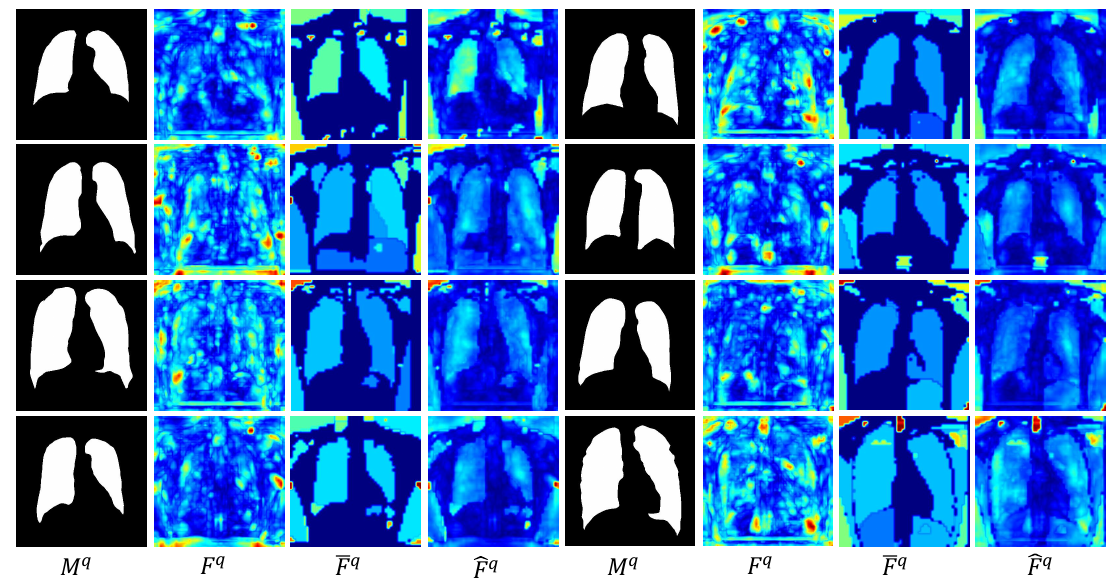}
\caption{Qualitative analysis results on Chest X-Ray: $M^q$ represent the query ground truth mask. $F^q$, $\bar{F}^q$, $\hat{F}^q$ correspond to the feature map extracted by the backbone network, the 3D feature map of the visual prompts, and the feature map adapted to the new task, respectively.}
\label{vis7}
\end{figure*}

\begin{figure*}[t!]
\centering
\includegraphics[width=1.0\linewidth]{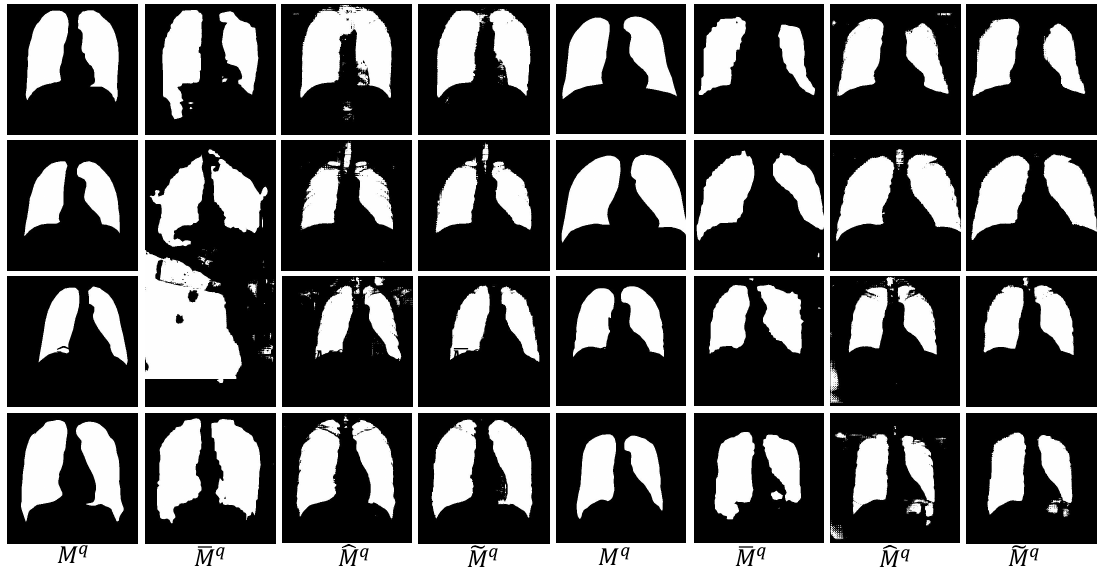}
\caption{Qualitative analysis results Chest X-Ray: $M^q$ represent the query ground truth mask. $\bar{M}^q$, $\hat{M}^q$ and $\Tilde{M}^q$ represent the model’s prediction,
SAM’s prediction, and the final segmentation result after refinement, respectively.}
\label{vis8}
\end{figure*}

\begin{figure*}[t!]
\centering
\includegraphics[width=1.0\linewidth]{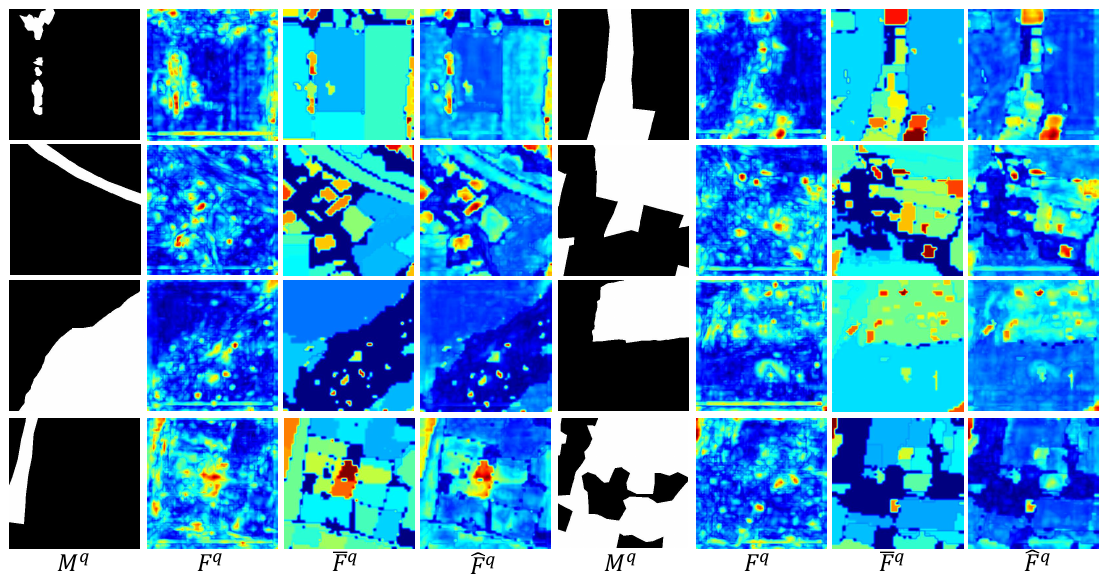}
\caption{Qualitative analysis results on Deepglobe: $M^q$ represent the query ground truth mask. $F^q$, $\bar{F}^q$, $\hat{F}^q$ correspond to the feature map extracted by the backbone network, the 3D feature map of the visual prompts, and the feature map adapted to the new task, respectively.}
\label{vis9}
\end{figure*}

\begin{figure*}[t!]
\centering
\includegraphics[width=1.0\linewidth]{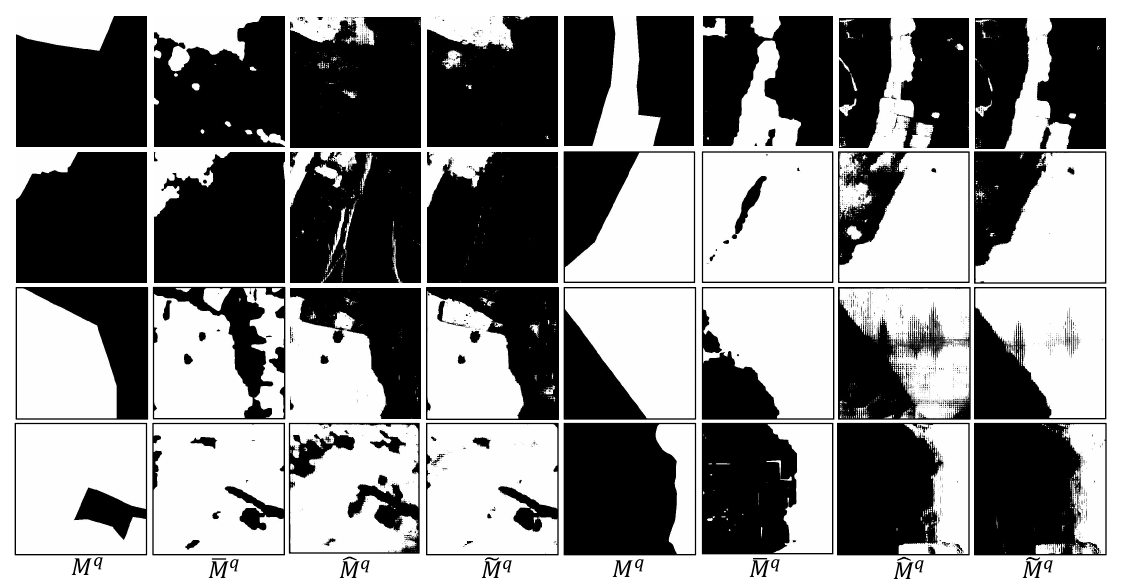}
\caption{Qualitative analysis results on Deepglobe: $M^q$ represent the query ground truth mask. $\bar{M}^q$, $\hat{M}^q$ and $\Tilde{M}^q$ represent the model’s prediction,
SAM’s prediction, and the final segmentation result after refinement, respectively.}
\label{vis10}
\end{figure*}

Visualization Understanding of the Proposed Modules in GPRN. As shown in Fig. \ref{vis3} to \ref{vis10}, we provide two sets of visualizations for each target dataset. One set compares the original feature $F^q$ extracted by the backbone with the new feature obtained by adding the 3D visual prompts feature map $\bar{F}^q$ to the original feature. The other set compares the initial prediction after fusion with SAM predictions against the initial prediction. The results indicate that our proposed modules are significantly effective across all datasets.

\begin{table}[h]
	\setlength{\abovecaptionskip}{5pt}
	\setlength{\belowcaptionskip}{10pt}
	\caption{Performance of GPRN with and without eliminating the overlapping regions of masks.}
	\renewcommand\arraystretch{1.2}
	\centering
	\resizebox{0.9\linewidth}{!}{
		\begin{tabular}{ccccccc} 
			\toprule
			Setting & Deepglobe & ISIC & Chest X-Ray & FSS-1000   \\ 
			\hline
			w(ours)         &    51.33   &   66.42    &   79.93    &   78.20      \\
			w/o  &  51.26     &   65.98    &   77.79    &   78.22           \\
			\bottomrule
	\end{tabular}}
	\label{tab3}
\end{table}

\end{document}